\pdfoutput=1

\documentclass[11pt]{article}

\usepackage[final]{acl}

\usepackage{times}
\usepackage{latexsym}
\usepackage{booktabs}
\usepackage{amssymb}
\usepackage{tabularx}
\usepackage{makecell}
\usepackage{float}
\usepackage{amsmath}
\usepackage{amssymb}
\usepackage{multirow}
\usepackage{pgfplots}
\usepackage{hyperref}

\DeclareMathOperator*{\argmin}{arg\,min}
\newcommand\norm[1]{\left\lVert#1\right\rVert}

\pgfplotsset{
    compat=1.16,
    tick label style={font=\small},
    label style={font=\small},
    legend style={font=\small},
    title style={font=\normalsize},
}
\usepackage[T1]{fontenc}

\usepackage[utf8]{inputenc}

\usepackage{microtype}

\usepackage{inconsolata}

\usepackage{graphicx}

\usepackage{algorithm}
\usepackage{algpseudocode}
\usepackage{makecell}
\usepackage{pgfplots}
\usepgfplotslibrary{statistics}
\usepackage{float} 
\usepackage{caption} 
\usepackage{lipsum} 
\usepackage{subcaption} 
\title{Generalizable and Stable Finetuning of Pretrained Language Models on Low-Resource Texts}

\author{
    Sai Ashish Somayajula \quad Youwei Liang \quad Li Zhang \quad Abhishek Singh \quad Pengtao Xie \\
    UC San Diego, USA\\
    \texttt{\{ssomayaj, p1xie\}@ucsd.edu}
}

\begin{document}
\maketitle
\begin{abstract}
Pretrained Language Models (PLMs) have advanced Natural Language Processing (NLP) tasks significantly, but finetuning PLMs on low-resource datasets poses significant challenges such as instability and overfitting. 
Previous methods tackle these issues by finetuning a strategically chosen subnetwork on a downstream task, while keeping the remaining weights fixed to the pretrained weights. However, they rely on a suboptimal criteria for sub-network selection, leading to suboptimal solutions. 
To address these limitations, we propose a regularization method based on attention-guided weight mixup for finetuning PLMs. Our approach represents each network weight as a mixup of task-specific weight and pretrained weight, controlled by a learnable attention parameter, providing finer control over sub-network selection. Furthermore, we employ a bi-level optimization (BLO) based framework on two separate splits of the training dataset, improving generalization and combating overfitting. We validate the efficacy of our proposed method through extensive experiments, demonstrating its superiority over previous methods, particularly in the context of finetuning PLMs on low-resource datasets.~\footnote{Our code is available at \href{https://github.com/Sai-Ashish/Attention_guided_weight_mixup_BLO}{https://github.com/Sai-Ashish/Attention\_guided\_weight\_mixup\_BLO}.}
\end{abstract}

\section{Introduction}

Pretraining large language models on a large corpus of unlabeled texts and further finetuning them on downstream tasks have been a common practice in natural language processing (NLP), resulting in significant advances in tasks such as sentiment classification, natural language inference, and text generation \cite{devlin2018bert, liu2019roberta, lewis2019bart, raffel2020exploring}. 

However, conventional finetuning of pretrained language models (PLMs) presents several challenges. First, PLMs are prone to instability in finetuning, characterized by high variance in performance for different weight initializations even when using the same hyperparameters, especially on small datasets \cite{ziser2019task, devlin2018bert, phang2018sentence, lee2019mixout, dodge2020fine, zhang2020revisiting}.
Moreover, PLMs, due to their extremely large capacity, are prone to overfitting when finetuned on small downstream datasets, leading to poor generalization on test set~\cite{belinkov2020variational, aghajanyan2020better,kuang2021balance}. 
Consequently, adapting PLMs to a variety of 
low-resource tasks, while preserving stability and maximizing generalization, remains a significant challenge in the field.



Finetuning a strategically chosen sub-network on a downstream task, while keeping the remaining weights fixed to the pretrained weights, has effectively mitigated these 
challenges. 
Within this umbrella, ${\text{CHILD-TUNING}_D}$~\cite{xu2021raise} and DPS dense~\cite{zhang2022fine} are promising. In ${\text{CHILD-TUNING}_D}$ \cite{xu2021raise}, a static sub-network, termed ``child network'',
is first selected based on the Fisher Information Matrix (FIM) and this child network is subsequently updated during finetuning. Dynamic Parameter Selection (DPS) \cite{zhang2022fine} further refines this approach by dynamically selecting the child network during PLM finetuning using FIM as a guiding principle.

Nevertheless, these prior works exhibit certain limitations. 
FIM, which is empirically calculated using the training dataset to identify important network parameters, may not be optimal for child network selection, especially in low-resource settings where data scarcity can skew the gradient that is used to compute FIM~\cite{kunstner2019limitations}. This scenario can lead to unimportant parameters being subsequently chosen in the sub-network, deteriorating performance on downstream tasks. Moreover, \citet{soen2021variance} theoretically shows that empirically determined FIM deviates significantly from the true FIM when the number of samples is low.
Thus, the discrete selection of child networks based on heuristics (i.e., FIM), may result in selection of suboptimal child networks for downstream tasks.
These limitations necessitate a departure from FIM-based discrete child network selection strategies in favor of one that selects a child network based on the model's downstream task performance. Consequently, we advocate for a continuous optimization approach for child network selection that does not rely on FIM, guided by the goal of optimizing performance in downstream tasks.

In this work, we propose an end-to-end framework that converts prior heuristic-based approaches to discrete child network selection into a continuous relaxation that can be optimized using gradient descent. The crux of our method is an attention-guided weight mixup mechanism that facilitates this transformation. Each weight is represented as a weighted sum of task weights (from downstream task finetuning) and pretrained weights (which is frozen), controlled by a mixing coefficient, which we refer to as ``attention parameter''~\footnote{This attention parameter, $\alpha$, is \emph{not} to be confused with the attention layers of the transformers.}. This setup leads to a continuous relaxation: a larger attention value of a weight indicates that it is more likely to belong to the child network and vice versa. We formulate the learning of task weights and the attention parameters in a bi-level optimization (BLO) framework~\cite{feurer2015initializing} on two different splits of the training dataset. 
In the lower level of our formulation, we update task weights by minimizing loss on the first split, while in the upper level of the framework, we update the attention parameter by minimizing loss on the second split. This bi-level learning of task weights and the attention parameters on two different splits of the training dataset sidesteps the FIM-based heuristic, ensuring it is more adapted to the downstream task, leading to better performance (Sec.~\ref{sec:exp}).

\noindent Our major contributions are summarized below:
\begin{itemize}
    \item We address the crucial issue of finetuning PLMs on low-resource datasets by leveraging an attention-guided weights mixup strategy. In this approach, each weight is represented as a mixup of task weights and pretrained weights, controlled by an attention parameter. This method is a continuous relaxation of the prior discrete sub-network selection approach.
    \item 
    We capitalize on an inter-dependency between task weights and attention parameter to formulate the learning objective as a BLO problem, which allows us to learn the attention parameter and model weights on two separate splits of the training set respectively. This has been a key of our method to combating overfitting and increasing stability. 
    \item We extensively evaluate our method on several datasets of the GLUE benchmark in low-resource settings, demonstrating improvements over several baselines. Our method has also achieved enhanced stability over standard finetuning across different PLMs.
\end{itemize}

\section{Related works}

\subsection{Generalizable finetuning}
Finetuning PLMs on small downstream datasets can lead to overfitting and instability. Several techniques have been proposed to address this issue. Weight decay \cite{daume2009frustratingly} incorporates a regularization term with a fixed trade-off coefficient to mitigate overfitting, while RecAdam \cite{chen2020recall} improves weight decay by introducing a time-varying trade-off coefficient for regularization loss, and an L2 loss between task and pretrained weights, which is used to regularize the finetuning. Top-K-layer finetuning \cite{pmlr-v97-houlsby19a} focuses on updating the weights of the top $K$ layers and keeping the pretrained bottom layer weights intact, thereby regularizing the finetuning. R3F \cite{aghajanyan2020better} introduces parametric noise into input sentence embeddings for better generalization. R-Dropout~\cite{wu2021r} minimizes the Kullback-Leibler divergence of the predictions from two sub-models created by different dropouts, thereby fostering prediction consistency. Re-init \cite{zhang2020revisiting} reinitializes the pooler and top $K$ Transformer layers before finetuning BERT, which is found to perform better than vanilla finetuning. Mixout~\cite{lee2019mixout} proposes to randomly replace task weights
with their corresponding pretrained weights to mitigate overfitting and improve stability. 

Instead of randomly replacing some task weights, sub-network optimization methods such as ${\text{CHILD-TUNING}_D}$ \cite{xu2021raise} and DPS dense \cite{zhang2022fine} leverage FIM to select a subset of the model parameters that are relevant to downstream tasks to finetune, resulting in better performance than Mixout and other methods above. 
However, FIM-based methods have some limitations especially in low-resource scenarios.
To address these issues, we propose to use the model performance on the downstream task to select the sub-network to finetune. To overcome overfitting, we propose to update the task weights on a training set and use the model performance on a separate validation set for the sub-network selection, resulting in a BLO framework. 
Appendix~\ref{sec:appendix C.} has more details about each method.

\begin{figure*}[ht]
\centering
\includegraphics[scale=0.7]{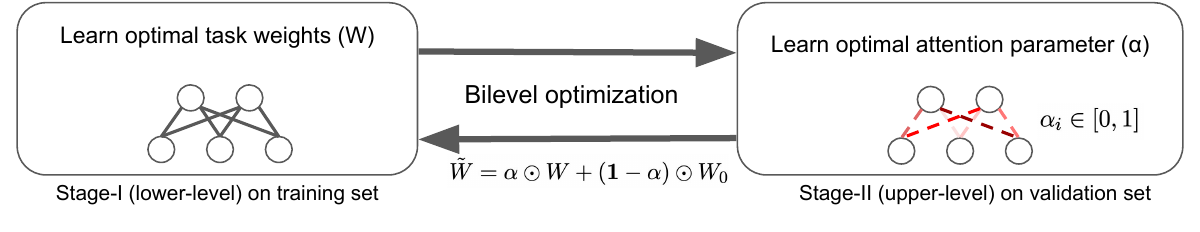}
\caption{An overview of our proposed method: learning the task weights $W$ and the attention parameter $\alpha$ in a bilevel optimization framework. The final network weight $\tilde{W}$ is a combination of the pretrained weight $W_0$ and the task weight $W$ via the learned attention parameter $\alpha$.}
\label{framework-fig}
\end{figure*}

\subsection{Bi-level optimization (BLO)}

BLO refers to a class of optimization problems in which one optimization problem (lower level) is nested within another optimization problem (upper level)~\cite{sinha2017review}. Many problems in machine learning, including neural architecture search \cite{liu2018darts}, hyperparameter tuning \cite{feurer2015initializing, baydin2017online}, data selection \cite{shu2019meta, wang2020meta, ren2020not}, meta-learning \cite{finn2017model}, and noisy label correction \cite{baydin2017online}, can be \emph{formulated} as a BLO problem. In these applications, the upper-level optimization is responsible for learning meta-variables, such as hyperparameters and architectures, by minimizing the validation loss, while the lower-level optimization focuses on learning model weights by minimizing the training loss. 
Although these works, as well as ours, are formulated as BLO problems, our contributions are notably different since we target different domains. Recognizing the limitations of existing sub-network optimization methods, we introduce a continuous relaxation approach that utilizes an attention-guided weight mixup strategy. Subsequently, we suggest employing BLO, capitalizing on the problem's structure as detailed in Section~\ref{methods}.
\section{Method}
\label{methods}
We introduce ``Attention-Guided Weights Mixup'', depicted in Fig~\ref{framework-fig}, an approach to improving stability and performance on small downstream datasets in PLMs. Central to our approach is a unique representation of weights as a blend of the pretrained weights (those prior to finetuning) and the task weights (post-downstream task finetuning), modulated by an ``attention parameter''. This coefficient represents the degree of emphasis or ``attention'' to be placed on a pretrained weight during the computation of the resultant weight. To mitigate overfitting, we propose to learn the attention parameter and the task weights using BLO on different splits of the training dataset; pretrained weights are frozen. This initial phase, termed search phase, aims to find optimal attention parameters, i.e., to find the optimal child network. Subsequently, in the finetune phase, the task weights are further adjusted using the entire training dataset and the learned attention parameters. 

\subsection{Continuous relaxation of child network selection}

To overcome the limitations of FIM-based discrete child network selection methods, we introduce a unique continuous relaxation approach. This approach is based on an attention-guided weight mixup mechanism.
To elaborate on this formulation, consider
the task weights and the pretrained weights of a PLM, denoted by $W \in \mathcal{R}^{N \times M}$ and $W_0 \in \mathcal{R}^{N \times M}$ respectively, where $N$ and $M$ are the dimensions of the weight. An attention parameter, denoted by $\alpha \in {[0,1]}^{N \times M}$, is associated with each pretrained weight. This parameter indicates the relative importance of the pretrained weight compared to the task weight. We compute the resultant weight, $\Tilde{W}$. This weight is represented as an interpolation between the task and pretrained weights, the balance of which is dictated by the respective attention parameter:
\[\Tilde{W} = g(W, \alpha, W_0) = \alpha \odot W + (\mathbf{1} - \alpha) \odot W_0 \]
where, $\odot$ denotes the element-wise multiplication operation, and $\mathbf{1} \in \mathcal{R}^{N \times M}$ denotes the matrix with all its entries 1's. The discrete child network selection can be perceived as a special case within this formulation. To elaborate, if $\alpha$ equals $\mathbf{1}$, the implication is that the weight belongs to the child network. Conversely, if $\alpha$ is $\mathbf{0}$ (the matrix with all its entries 0's), the weight is tied to the pretrained, non-child network. Nonetheless, the entries of $\alpha$ are not restricted to the extremes of 0 or 1; instead, they can assume any continuous value within this range. If the learned attention parameter, $\alpha$, leans closer to $\mathbf{1}$, it signals a predominant influence from the task weight compared to the pretrained counterpart and vice versa. Thus, $\alpha$ serves as a mechanism that allows a flexible transition from the discrete child network selection to a continuous scale. In particular, $\alpha$ is designed to regulate the balance between the pretrained and task weights in the computation of the resultant weight.
\subsection{Learning task weight and attention parameter via bi-level optimization}

With continuous relaxation, the task of child network selection without using FIM morphs into determining attention parameters. 
Consequently, the task weights become dependent on these attention parameters, i.e. the chosen child network. However, in a reciprocal relationship, task weights should be considered while learning the attention parameters.
This is because the attention parameters aim to ascertain an optimal blend of pretrained and task weights in the resultant weight computation, engendering a mutual dependency. 

Navigating this intricate interdependency calls for a nuanced approach. Our approach to optimizing task weights $W$ and attention parameters $\alpha$ leverages a BLO framework composed of two interdependent learning stages. In the first stage, the finetuning of task weights occurs, driven by the minimization of the training loss. 
Subsequently, in the second stage, the attention weights are updated to minimize the model's validation loss.
Given the optimal task weights $W^*$ on the training set in the first stage, the goal of BLO is to learn the optimal attention parameters $\alpha^*$ that determines the right mix of $W^*$ and $W_0$ in the resultant weight estimation by minimizing the validation loss. 

To facilitate this, we partition the original training dataset, represented as $\mathcal{D}^{\text{tr}}$, into two subsets of 80\% and 20\% split. 50\%-50\% split setting was explored however 80\%-20\% split gave better empirical results (more insights in Appendix~\ref{sec:appendix D.}). 
The first subset, the BLO training dataset ($\mathcal{D}^{\text{B-tr}}$), is utilized in the first stage. The second subset, the BLO validation dataset ($\mathcal{D}^{\text{B-val}}$), is used in the second stage. While this two-stage optimization process unfolds, we freeze the pretrained weights. By doing so, our approach provides finer control over network optimization and capitalizes on the potential of pretrained weights. This phase to learn the parameter importance and the model parameters is termed the ``search phase.'' This phase is followed by the ``finetune phase'' that further learns $W$ with the learned $\alpha$ fixed on the entire training dataset.

\paragraph{Stage I - learning \texorpdfstring{$W$}{W}}
In this stage, we solve for the optimal task weights on BLO training dataset $\mathcal{D}^{\text{B-tr}}$. The task weights $W$ are learned by minimizing the training loss. As explained above, each weight of the PLM is represented as an interpolation of the task weight $W$ and the pretrained weight $W_0$ weighted by an attention parameter $\alpha, \alpha_{ij} \in [0,1], \forall i,j$. 
The following optimization problem is solved at this level,

\small
\begin{equation}
\label{eq:sum_loss1}
 W^{*}(\alpha) =  \argmin_{W} \mathcal{L}(g(W, \alpha, W_0); \mathcal{D}^{\text{B-tr}}) + \lambda_1 \norm{W}^2_F
\end{equation}
\normalsize
\noindent where $\mathcal{L}$($\cdot$) is the cross-entropy loss, $\alpha$ denotes the attention parameter, $\lambda_1$ is the weight decay of $W$, and $\norm{\cdot}_F$ is the Frobenius norm. Weight decay is commonly used while finetuning PLMs and we too introduce a weight decay term on the task weights $W$. The optimal task weights, denoted by $W^{*}(\alpha)$, are a function of $\alpha$. This dependency is enforced because the optimal weights are learned by minimizing the loss in Equation~\ref{eq:sum_loss1} on $\mathcal{D}^{\text{B-tr}}$, which is a function of $\alpha$. The attention parameter $\alpha$ is not learned in this stage else a complex solution will be learned that overfits the BLO training dataset.

\paragraph{Stage II - learning \texorpdfstring{$\alpha$}{alpha}}
In this stage, the attention parameters are learned on the BLO validation dataset $\mathcal{D}^{\text{B-val}}$, given the optimal task weights learned in the previous stage. The model $W^*(\alpha)$ is evaluated on $\mathcal{D}^{\text{B-val}}$. The model's validation loss is a function of $\alpha$. The attention parameters are learned by minimizing the validation loss (a function of $\alpha$). The following optimization problem is solved at this stage,

\small
\begin{equation}
\label{eq:validation_stage}
 \min_{\alpha} \quad \mathcal{L}(g(W^*(\alpha), \alpha, W_0); \mathcal{D}^{\text{B-val}}) + \lambda_2 \norm{\alpha}^2_F 
\end{equation}
\normalsize
\noindent where $\lambda_2$ is the weight decay of $\alpha$. Weight decay on $\alpha$ encourages the values of $\alpha$ to be close to $\mathbf{0}$ encouraging a higher contribution from the pretrained weights in the resultant weight estimation.

\paragraph{Low-rank approximation of \texorpdfstring{$\alpha$}{alpha}}
Since we mainly target the low-resource domain, 
we use a low-rank approximation of $\alpha \in {[0,1]}^{N \times M}$ to mitigate overfitting
~\cite{hu2021lora}. Specifically, we express the $\alpha$ matrix as the product of two matrices with lower ranks (of rank $r$):
\begin{align}
    \alpha = \frac{1}{r} \odot \mathcal{F}(\alpha_1, \alpha_2) \label{eq:alpha_low_rank}
\end{align}
where $\mathcal{F}(\cdot, \cdot)$ denotes matrix multiplication, and both $\alpha_1 \in \mathbb{R}^{N \times r}$ and $\alpha_2 \in \mathbb{R}^{r \times M}$ have a rank of $r$. 
We restrict ${\{\alpha_1\}}_{ij} \in [0,1], \forall i,j$ and ${\{\alpha_2\}}_{ij} \in [0,1], \forall i,j$. The normalization constant $r$ in Eq.~\ref{eq:alpha_low_rank} can ensure the entries of $\alpha$ are in [0,1]. We use the following weight formulation that proves to have better performance empirically after rank decomposition:

\small
\[g(W, \alpha, W_0) = \mathcal{F}(\alpha_1, \alpha_2) \odot W + \mathcal{F}(1-\alpha_1, 1-\alpha_2) \odot W_0 \]
\normalsize


\paragraph{Our bi-level optimization framework}
Following BLO is formulated,
\begin{equation}
\label{eq:overall}
\small
\begin{aligned}
  \min_{\alpha} \quad & \mathcal{L}(g(W^*(\alpha), \alpha, W_0); \mathcal{D}^{\text{B-val}}) + \lambda_2 \norm{\alpha}^2_F \\
 \text{s.t.} \quad & \begin{aligned}[t]
    W^{*}(\alpha) &=  \arg\min_{W} \left[ \mathcal{L}(g(W, \alpha, W_0); \mathcal{D}^{\text{B-tr}}) \right. \\
    & \left. + \lambda_1 \norm{W}^2_F \right]
  \end{aligned}
\end{aligned}
\end{equation}
\normalsize
\noindent There are two optimization problems each corresponding to a learning stage. From bottom to top, each learning stage corresponds to stage I and stage II that are dependent on each other via the loss function. Both learning stages are performed in an end-to-end fashion. The optimal task weights $W^*(\alpha)$ are learned by minimizing the loss function defined on $\mathcal{D}^{\text{B-tr}}$. The optimal task weights are a function of attention parameters $\alpha$ because the training loss is dependent on $\alpha$ via the resultant weight definition defined above. In the second stage, attention parameters $\alpha$ are learned by minimizing the validation loss on $\mathcal{D}^{\text{B-val}}$. This would influence the training loss in stage I, which influences the solution $W^*(\alpha)$.

\paragraph{Optimization algorithm}
Our approach consists of two phases, shown in Algorithm~\ref{optim_algorithm}. (I) Search phase: The optimal attention parameters are estimated in this phase posed as a BLO~(\ref{eq:overall}). 
We calculate the gradient of Eq.\ref{eq:sum_loss1} with respect to $W$ and approximately update $W^*(\alpha)$ using one-step gradient descent. Similarly, we update $\alpha$ using one-step gradient descent of Eq.\ref{eq:validation_stage}; however, we encounter a hessian-vector product while estimating the gradient due to the chain rule, which is computationally expensive to calculate. This is approximated using finite-difference approximation~\cite{liu2018darts}, as described in detail in Appendix~\ref{sec:appendix A.}.
The algorithm learns $\alpha^{\prime} \approx \alpha^{*}$. (II) Finetune phase: In this phase, we further finetune the task weights on entire training dataset ($\mathcal{D}^{\text{tr}}$) using the estimated attention parameters $\alpha^{\prime}$.

\section{Experiments}
\label{sec:exp}

\subsection{Datasets}
We conduct experiments on various datasets from the GLUE benchmark~\cite{warstadt2018neural, wang-etal-2018-glue} following \citet{xu2021raise, zhang2022fine}. The datasets we have chosen for our evaluation cover an extensive range of linguistic tasks. These include sentiment classification (SST-2), natural language inference (RTE, QNLI, MNLI), paraphrasing and similarity assessment (MRPC, STS-B, QQP), and finally, the evaluation of linguistic acceptability (CoLA). We finetune the models on the training set for each of the mentioned datasets and evaluated its performance on the original development set using the checkpoint obtained at the end of training following~\citet{xu2021raise, zhang2022fine}. More information about each of the chosen datasets can be found in Appendix~\ref{sec:appendix B.}.

\begin{table*}[t]
\centering 
\small
\begin{tabular}{c | c c c c} 
\toprule
\multicolumn{1}{c|}{\textbf{Training split}} & \textbf{Vanilla} & \textbf{$\text{CHILD-TUNING}_D$} & \textbf{DPS Dense} & \textbf{Ours}\\ \midrule
\textbf{300} & 62.54 $\pm$ 6.57  & 62.47 $\pm$ 5.5 & 61.69 $\pm$ 5.62  & \textbf{68.97} $\pm$ 3.09 \\
\textbf{500} & 65.85 $\pm$ 4.57 & 68.35 $\pm$ 4.36 & {{68.99}} $\pm$ {{2.92}} & \textbf{72.42} $\pm$ 2.14  \\
\textbf{1000} & 73.19 $\pm$ 2.62 & 74.07 $\pm$ 2.75 & {{75.00 $\pm$ 1.61}} & \textbf{76.68} $\pm$ 1.58\\
\bottomrule
\end{tabular}%

\caption{
We compare our method with Vanilla, $\text{CHILD-TUNING}_D$, and DPS dense method using $\text{BERT}_{\text{LARGE}}$~\cite{lee2019mixout} across 300, 500, and 1000 training data splits. Reported results are the averaged evaluation metrics over all eight GLUE datasets for each training data split. Due to space constraints, detailed results for each of the eight datasets can be found in Table~\ref{tab:low-resource-300-500-1000}. The highest performance in each row is indicated in \textbf{bold}. 
}
\label{tab:low-resource}
\end{table*}

\subsection{Experimental setup}\label{section:experimental_setup}
We use the PLM codes provided by Huggingface~\cite{wolf-etal-2020-transformers} and follow their default settings unless specifically mentioned. Rank of $\alpha$, $r$ is set to 1 in this work. More detailed information about the hyperparameters settings in the search and the finetune phase, such as batch size, training steps, for $\text{BERT}_{\text{LARGE}}$~\cite{devlin2018bert}, $\text{RoBERTa}_{\text{LARGE}}$~\cite{liu2019roberta}, $\text{BART}_{\text{LARGE}}$~\cite{lewis2019bart}, $\text{DeBERTa}_{\text{LARGE}}$~\cite{he2020deberta}, and $\text{XLNet}_{\text{LARGE}}$~\cite{yang2019xlnet} can be found in Appendix~\ref{sec:appendix D.}. The averaged results over ten random seeds 
are reported in the paper.
In this work, we primarily compare with ${\text{CHILD-TUNING}_D}$~\cite{xu2021raise} and DPS dense~\cite{zhang2022fine}, as these methods share a similar motivation to ours:
improving performance on small downstream tasks by finetuning on a strategically chosen sub-network. 

\subsection{Performance on low-resource scenarios}\label{sec:low_resource}

\begin{figure}[t]
    \centering
    \includegraphics[width=\linewidth]{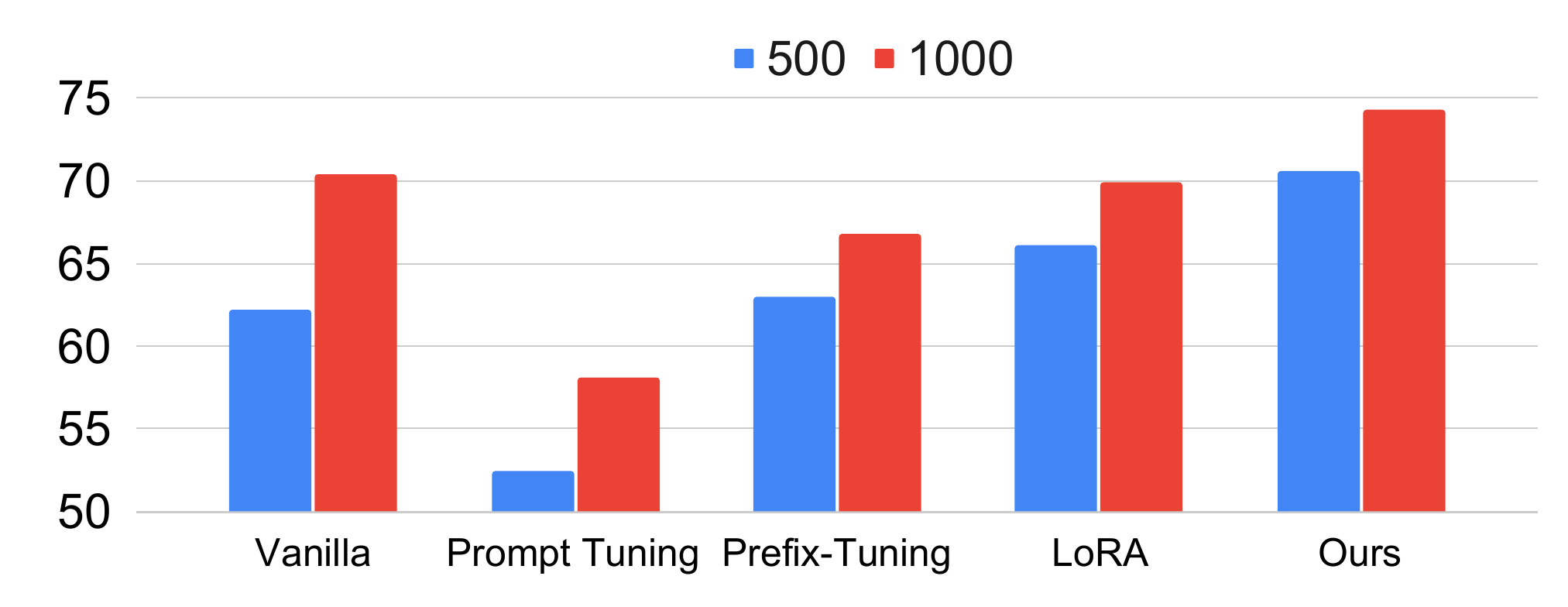}
    \caption{
    Averaged performance across CoLA, RTE, STSB, and MRPC datasets for Vanilla, Prompt Tuning, Prefix-Tuning, LoRA, and our method in low-resource scenarios with 500 and 1000 training instances. Results on each dataset are presented in Table~\ref{tab:lora}.
    }
    \label{fig:peft_fig}
\end{figure}

In this experiment, we investigate the effectiveness of our proposed method for finetuning PLMs on extremely small datasets. Specifically, we downsample 8 GLUE datasets by randomly selecting 300, 500 and 1K training examples following~\citet{zhang2022fine}.~\footnote{Consistent with prior methods~\cite{xu2021raise,zhang2022fine}, a random subset 
is chosen from entire training dataset based on a seed, to avoid bias towards any specific subset. Evaluation is performed over ten random seeds.
} Table~\ref{tab:low-resource} summarizes our results. Comparing our approach to DPS dense and ${\text{CHILD-TUNING}_D}$, we observe substantial improvements in average scores.
On average, our method outperforms the best baseline by 6.43\%, 3.43\%, and 1.68\% on 300, 500, and 1K training samples scenarios, respectively. 

The superior performance of our method can be attributed to two key designs in our framework. First, our method employs continuous weight mixup, assigning attention parameters to each pretrained weight. This allows for a more adaptable and dynamic determination of the importance of pretrained weight during finetuning. 
Second, our BLO framework strengthens our method by optimizing task weights and attention parameters on two distinct splits of the training dataset. This strategy effectively combats overfitting and ensures that attention parameters are learned based on validation performance, leading to improved generalization on unseen test data. To summarize, our method shows strong potential for improving performance on low-resource NLP tasks.

\begin{table*}[t]
\centering 
\small
\begin{tabular}{l | c | c c c c c c c c l l} 
\toprule
\textbf{Models} & Methods & \multicolumn{2}{c}{\textbf{CoLA}} & \multicolumn{2}{c}{\textbf{MRPC}} & \multicolumn{2}{c}{\textbf{RTE}} & \multicolumn{2}{c}{\textbf{STSB}} & \multicolumn{2}{c}{\makebox[5em][l]{\textbf{Average}}} \\
\cline{3-12} & & Mean & Std & Mean & Std & Mean & Std & Mean & Std & Mean & Std\\
\hline
\multirow{2}{*}{BERT} & Vanilla & 64.11 & {1.33}  & 90.80 & 1.77  & 70.69 & 2.83  & 89.92 & 0.61  & 78.88 & 1.64 \\
& Ours & \textbf{66.07} & {1.35} & \textbf{91.84} & {0.37}  & \textbf{73.43} & {1.52} & \textbf{90.34} & {0.48}  & $\textbf{\text{80.42}}({\text{\textcolor{blue}{+1.54}}})$ & ${\text{0.93}} ({\text{\textcolor{blue}{-0.71}}})$\\
\hline
\multirow{2}{*}{BART} & Vanilla &  58.54 & 1.41 & 92.03 & 0.73 & 81.84 & 1.41 & 91.54 & 0.40 & 80.99 & 0.99\\
& Ours & \textbf{60.15} & {0.81} & \textbf{92.33} & {0.40}  & \textbf{84.26} & {0.54} & \textbf{92.20} & {0.09} & $\textbf{\text{82.23}}({\text{\textcolor{blue}{+1.24}}})$ & ${\text{0.46}}({\text{\textcolor{blue}{-0.53}}})$\\
\hline
\multirow{2}{*}{RoBERTa} & Vanilla &  66.06 & 2.07 & 92.25 & 0.57 & \underline{78.52} & \underline{13.01} & 91.89 & 0.31 & 82.18 & 3.99\\
& Ours & \textbf{66.52} & {1.45} &  \textbf{92.58} & {0.48} & \textbf{84.22} & {1.44} & \textbf{92.21} & {0.08} & $\textbf{\text{83.88}}({\text{\textcolor{blue}{+1.70}}})$ & ${\text{0.86}}({\text{\textcolor{blue}{-3.13}}})$\\
\hline
\multirow{2}{*}{DeBERTa} & Vanilla &  63.74 & 1.34 & 92.31 & 0.37 & 85.59 & 1.58 & 91.74 & 0.17 & 83.34 & 0.86\\
& Ours &  \textbf{65.96} & {1.15} & \textbf{92.32} & {0.28} &  \textbf{86.17} &  {1.47} &  \textbf{91.99} & {0.15} & $\textbf{\text{84.11}}({\text{\textcolor{blue}{+0.77}}})$ & $\text{0.76}({\text{\textcolor{blue}{-0.10}}})$\\
\hline
\multirow{2}{*}{XLNet} & Vanilla & \underline{40.93} & \underline{27.28} & 91.83 & 0.91 & \underline{71.17} & \underline{14.40} & 91.68 & 0.19 & 73.90 & 10.69 \\
& Ours & \textbf{61.66} & {1.95} & \textbf{92.19} & {0.38} &  \textbf{83.54} &  {1.44} &  \textbf{92.12} & {0.08} & $\textbf{\text{82.38}}({\text{\textcolor{blue}{+8.48}}})$ & ${\text{0.96}}({\text{\textcolor{blue}{-9.73}}})$ \\ 
\bottomrule
\end{tabular}

\caption{Comparison of our method and vanilla finetuning on five popular PLMs. We evaluated the models using ten runs with different random seeds and reported the results in terms of mean and standard deviation. Average score represents the average performance across four datasets, and the best scores are highlighted in \textbf{bold}. The \underline{underlined} values indicate occurrences of degenerate seeds.
}
\label{tab:pretrained_different}
\end{table*}

\begin{table*}[t]
\centering 
\small
\begin{tabular}{l c c c c c c c c c c} 
\toprule
\textbf{Methods} & \multicolumn{2}{c}{\textbf{CoLA}} & \multicolumn{2}{c}{\textbf{MRPC}} & \multicolumn{2}{c}{\textbf{RTE}} & \multicolumn{2}{c}{\textbf{STSB}} & \multicolumn{2}{c}{\textbf{Average}} \\
\cline{2-11} & Mean & Std & Mean & Std & Mean & Std & Mean & Std & Mean & Std\\
\hline
Vanilla & 64.11 & 1.33  & 90.80 & 1.77  & 70.69 & 2.83  & 89.92 & 0.61  & 78.88 & 1.64  \\
Mixout      & 64.42 & 1.51  & 91.31 & 1.08  & 72.05 & 1.67  & 90.39 & 0.57  & 79.54 & 1.21  \\
R3F         & 64.62 & 1.38  & 91.63 & 0.93 & 70.75 & 1.76  & 89.92 & 0.61  & 79.23 & 1.17  \\
R-Dropout   & 64.14 & 1.58  & \textbf{91.87} & {{0.78}} & 70.24 & 2.83  & 90.25 & 0.49  & 79.13 & 1.42  \\
$\text{CHILD-TUNING}_D$      & 64.85 & {{1.32}}  & 91.52 & 0.81  & 71.69 & 1.95  & 90.42 & {{0.44}} & 79.62 & 1.13  \\
Re-init     & 64.24 & 2.03  & 91.61 & 0.80  & 72.44 & 1.74  & \textbf{90.71} & {0.14} & 79.75 & 1.18  \\
DPS Dense   & {{64.98}} & {1.08}  & 91.50 & 0.83  & {{73.14}} & 1.97  & {{90.51}} & 0.55  & {{80.03}} & {{1.11}}  \\
DPS Dense (Our run) & 64.08 & 1.50 & 90.25 & 2.21  & 71.92 & {1.45} & 90.20 & 0.47  & {79.11} & {1.41} \\
\midrule
Ours & \textbf{66.07} & {1.35} & {{91.84}} & {0.37}  & \textbf{73.43} & {{{1.52}}} & 90.34 & 0.48  & \textbf{80.42} & {0.93} \\
\bottomrule
\end{tabular}
\caption{Comparison of our method with other finetuning methods on four small datasets (CoLA, RTE, MRPC, STSB), known for causing instability in $\text{BERT}_{\text{LARGE}}$~\cite{lee2019mixout}. The baseline results are taken from the DPS dense paper~\cite{zhang2022fine}. The mean and standard deviation (std) of ten random seeds are reported for each method.
We utilize the vanilla results on STS-B for R3F since it cannot be applied to the regression task of STS-B. \textbf{Bold} indicates the best performance. Double-sided t-tests were performed between our method and the vanilla method. The p-values are less than 0.05, indicating statistically significant performance improvement over vanilla.}
\label{tab:mainresult}
\end{table*}

\subsection{Comparison with parameter efficient finetuning methods} \label{sec:peft}

In this section, we compare our method with popular parameter-efficient finetuning (PEFT) techniques, including Prompt Tuning~\cite{lester2021power}, Prefix-Tuning~\cite{li2021prefix}, and LoRA~\cite{hu2021lora}. Our experiments were conducted on MRPC, STSB, CoLA, and RTE datasets under two low-resource settings with 500 and 1,000 data points, respectively. 
As shown in Figure~\ref{fig:peft_fig}, our method consistently outperforms these baselines. Considering performance on each dataset (Table~\ref{tab:lora}), LoRA lags behind vanilla finetuning on all datasets except CoLA. Prefix-Tuning underperforms compared to both vanilla finetuning and LoRA in most scenarios. The deteriorated performance of Prefix-Tuning is also observed in other low-resource settings~\cite{hu2021lora}. Besides, Prompt Tuning falls behind vanilla finetuning on all datasets (Table~\ref{tab:lora}). On average, as depicted in Figure~\ref{fig:peft_fig}, Prompt Tuning trails vanilla finetuning by a significant margin. The low performance of Prompt Tuning is particularly evident in models with fewer than 10 billion parameters~ \cite{lester2021power}.


It is worth noting that the primary aim of these PEFT methods is not necessarily to outperform vanilla finetuning in low-resource settings. Instead, they seek to deliver competitive performance without intensive resource utilization. Such methods are especially valuable when finetuning large models like GPT-2~\cite{gpt2} (774M) and GPT-3~\cite{gpt3} (175B). They finetune a few additional parameters while preserving the primary model backbone. In contrast, our method draws inspiration from regularization techniques such as $\text{CHILD-TUNING}_D$ and DPS dense, 
specifically designed to reduce overfitting and improve performance in low-resource finetuning scenarios. Our proposed method further improves the performance over these prior methods by addressing the challenges associated with FIM-based discrete sub-network selection by employing continuous optimization through attention-guided weight mixup.  

\subsection{Performance on different PLMs}\label{sec:diff_plms}

In this experiment, we investigate the effectiveness of our method when applied to various PLMs across different architectures compared to vanilla finetuning. We perform experiments using four GLUE tasks, chosen particularly due to their small training dataset sizes, on the language models mentioned in Section~\ref{section:experimental_setup}.
The results of these experiments are presented in Table~\ref{tab:pretrained_different}. 

Apart from BERT, our method consistently achieves better results on RoBERTa, which is trained on a larger pretraining corpus and employs more advanced self-supervised pretraining tasks. Notably, on the RTE dataset, the vanilla method yielded degenerate results for two random seeds, leading to poor performance and high standard deviation. This issue has been attributed to the vanishing gradients problem occurring for certain initialization~\cite{mosbach2020stability}. However, using the same initialization and settings, our method showed better results, and the degenerate results were not observed. Similar results were observed with XLNet. We observe a notable gain of 8.48\% over vanilla, along with a substantial decrease in the standard deviation by 9.73. Furthermore, we observe performance improvements on DeBERTa, a model with relative position encoding, and BART, an encoder-decoder-based model. These findings suggest that our method can effectively utilize the potential of pretrained weights, irrespective of the underlying model architecture, leading to enhanced finetuning performance across a variety of PLMs.
\subsection{A comparison of previous techniques on few-sample BERT finetuning}\label{sec:Main}
We compare our method to prior 
regularization-based approaches, namely Mixout~\cite{lee2019mixout}, R3F \cite{aghajanyan2020better}, R-Dropout~\cite{wu2021r}, Re-init~\cite{zhang2020revisiting}, ${\text{CHILD-TUNING}_D}$~\cite{xu2021raise}, and DPS dense~\cite{zhang2022fine}, following 
\citet{zhang2022fine, xu2021raise}. We specifically employ CoLA, RTE, MRPC, and STSB datasets for this evaluation, 
as finetuning $\text{BERT}_\text{LARGE}$ model on these small datasets cause instability 
\cite{lee2019mixout}. Table~\ref{tab:mainresult} summarizes our results. 

Our method surpasses all other baselines in terms of average scores, with a particularly notable improvement on the CoLA dataset over baselines, illustrating the effectiveness of our approach. Additionally, our method has the smallest standard deviation averaged over the datasets, indicating increased stability. In summary, our attention-guided weight mixup approach improves stability and performance on small datasets over baselines.
\subsection{Ablation studies}\label{sec:ablation_studies}

\paragraph{Joint-training} We conduct an ablation study contrasting our method with ``Joint Training'', a technique where the $W$ and $\alpha$ parameters are jointly optimized by minimizing the loss on the whole training set rather than using our proposed BLO framework. The results are summarized in Table~\ref{tab:average_values}. The results underscore the superior performance of our method across all datasets, yielding a higher average score and a reduced standard deviation. Notably, on the MRPC dataset (Table~\ref{tab:random_init}), Joint-Training performs worse than the vanilla baseline. This outcome suggests a potential pitfall of Joint-Training: the simultaneous learning of both $W$ and $\alpha$ parameters on the training dataset can lead to overfitting and result in a model that lacks generalizability on unseen test data. Conversely, our proposed approach, which uses a BLO framework to learn $W$ and $\alpha$ on two different splits of the training dataset, effectively mitigates overfitting, leading to better generalization on the test set, and enhancing overall performance.

\paragraph{Randomly fixed $\alpha$} 
We investigate the impact of randomly initializing the $\alpha$ parameters in the network and keeping them fixed throughout the optimization. We sample $\alpha_1$ and $\alpha_2$ parameters from a Gaussian distribution, ${\{\alpha_1\}}_{ij}, {\{\alpha_2\}}_{ij} \in \mathcal{N}(\mu_{\alpha}, \sigma_{\alpha})$, and fix them during finetuning, denoted as $\text{Random}_{\alpha}$. The Gaussian distribution has a mean of $\mu_{\alpha} = 1$ and standard deviations $\sigma_{\alpha}$ = \{0.005, 0.1, 0.45\}. After sampling from this distribution, we clip the $\alpha$ values to lie within the range of 0 to 1. The results in Table~\ref{tab:average_values} show that our method with learnable $\alpha$ outperforms the model with randomly initialized  $\alpha$, which underscores the importance of learning the attention parameters for child network selection. 

\begin{table}[t]
\centering 
\small
\begin{tabular}{l | c | c } 
\toprule
\textbf{Method} & \textbf{Mean} & \textbf{Std} \\
\midrule
Ours & \textbf{80.42} & 0.93 \\
Vanilla & 78.88 & 1.64 \\
Joint Training & 78.86 & 1.48 \\
\addlinespace 
\multicolumn{3}{l}{$\text{Random}_{\alpha}$} \\ 
\hspace{1em}$\sigma_{\alpha}$ = 0.005 & 79.36 & 1.03 \\
\hspace{1em}$\sigma_{\alpha}$ = 0.1 & 78.32 & 2.27 \\
\hspace{1em}$\sigma_{\alpha}$ = 0.45 & 69.29 & 5.44 \\
\bottomrule
\end{tabular}

\caption{Averaged performance across CoLA, RTE, STSB, and MRPC datasets for Vanilla, Joint-training and $\text{Random}_{\alpha}$ by varying $\sigma_{\alpha}$. Results on each dataset is presented in Table~\ref{tab:random_init}.}
\label{tab:average_values}
\end{table}

\begin{table*}[t]
\centering 
\small
\begin{tabular}{l | c | c | c | c | c | c} 
\toprule
\textbf{Method} & \textbf{Vanilla} & \textbf{R3F} & \textbf{R-Dropout} & \textbf{$\text{CHILD-TUNING}_D$} & \textbf{DPS} & \textbf{Ours} \\
\hline
\textbf{Time usage} & $\times$1 & $\times$1.64 & $\times$1.64 & $\times$3.13 & $\times$1.12 & $\times (1+0.8K)$ \\
\bottomrule
\end{tabular}
\caption{Comparison of time usage for different regularization based methods, where $K$ is a hyperparameter chosen using grid-search in $K = \{1,2,5\}$. Although we choose $K=5$ in all of our experiments, $K=2$ yields slightly worse but comparable performance and saves computational time.}
\label{tab:time_usage_comparison}
\end{table*}

\section{Computational costs and trade-offs} \label{sec:appendix E.}

Our method introduces attention-guided weights mixup through a BLO framework, leading to additional computational demands summarized in Table~\ref{tab:time_usage_comparison}, a characteristic shared with FIM-based strategies such as $\text{CHILD-TUNING}_D$ and DPS. Despite requiring comparable computation as current state-of-the-art techniques, our method offers performance improvements over several baselines, justifying the computational overhead. In terms of training efficiency, our method, in the worst-case scenario, requires a maximum of four times more training time than the vanilla method. However, in the best-case scenario, it is just 1.8 times the vanilla method. It is determined by the hyperparameter `$K$'. This increase is relatively small given the inherently smaller size of the low-resource datasets we focus on. On the other hand, it is pertinent to note that the inference time complexity remains consistent with other established methods. 

We compare the training costs of PEFT methods with our approach. Prompt Tuning necessitates approximately 0.2 times the training cost of vanilla fine-tuning, utilizing only 0.0067\% of the total parameters as trainable. In contrast, Prefix-Tuning and LoRA require about 0.4$\times$ and 0.3$\times$ the training cost of vanilla fine-tuning, respectively, with Prefix-Tuning utilizing 0.294\% and LoRA 0.236\% of the total parameters as trainable. For additional information, please refer to Appendix~\ref{sec:peft_appendix.}. Despite our method's higher training cost compared to the PEFT methods, as detailed in Section \ref{sec:peft}, it consistently surpasses vanilla fine-tuning in performance, whereas the PEFT methods generally fall short in most scenarios.

\section{Conclusions and future works}

In this work, we propose an attention-guided weight mixup mechanism to address issues in finetuning PLMs on low-resource datasets. Specifically, we represent each weight as a linear interpolation of the task weights and the pretrained weights, controlled by an attention parameter. Capitalizing on the inherent structure of this representation, we utilize a BLO framework to learn task weights and attention parameters on two different splits of training dataset. Our method demonstrated its effectiveness across several challenging datasets from the GLUE benchmark, outperforming baselines 
in low-resource scenarios. It also shows improved stability across different PLM architectures. Moreover, through ablation studies, we highlighted the importance of learning attention parameters continuously and the benefits of our BLO framework over the straightforward joint-training of model weights and attention parameters. 
We believe our method has the potential to be applied to other important areas, such as lifelong learning~\cite{parisi2019continual}. A common challenge in lifelong learning is retaining knowledge from previous tasks while adapting to new ones. Exploring our attention-guided weights mixup, optimized using BLO, for this application presents a promising avenue for future work.

\section{Limitations}\label{sec:limitations}

Our method improves the finetuning of PLMs but adds computational overhead, similar to ${\text{CHILD-TUNING}_D}$ and DPS dense, mainly due to calculating attention parameters in the BLO framework, akin to the computational demands of estimating and updating FIM in prior methods. However, its substantial performance gains over various baselines justify this extra cost.
In pursuit of minimizing this computational overhead, we contemplate an approach where the $\alpha$ parameters are intermittently updated, specifically every $T$ iterations. We believe this strategy will strike a balance between efficiency and performance. Delving into this modification poses an exciting avenue for subsequent research.
It would also be insightful to extend our method to multi-language tasks
to understand its adaptability and broader applicability in varying linguistic contexts.
\bibliography{anthology,custom}

\appendix

\section{Optimization algorithm} \label{sec:appendix A.}
To the best of our knowledge, our work is the first to employ BLO for selecting child network to prevent overfitting during the finetuning of large pretrained models. In this section, we dwell into the algorithm used to solve the proposed mathematical framework.
Various algorithms have been proposed to (approximately) solve a BLO problem. These can be broadly classified into two categories based on their approach to calculating upper-level gradients: implicit differentiation methods (such as Finite Difference~\cite{liu2018darts}, Neumann Series~\cite{lorraine2020optimizing}, and Conjugate Gradient~\cite{rajeswaran2019meta}) and iterative differentiation methods (including Reverse-mode Automatic Differentiation~\cite{finn2017model}). \citet{choe2022betty} empirically show that the Finite Difference method~\cite{liu2018darts} has the best performance among these methods.
Thus, we adopt this algorithm~\cite{liu2018darts} to solve our BLO~(\ref{eq:overall}) in the search phase. One-step gradient descent is used to approximate $W$.
\begin{equation}
\label{eqn:W}
\small
\begin{array}{ll}
\quad & W^{*}(\alpha) \approx W^{\prime} = W - \eta_{w}\nabla_{W} [\mathcal{L}(g(W, \alpha, W_0); \mathcal{D}^{\text{B-tr}}) \\& \quad \quad \quad \quad  + \lambda_1 \norm{W}^2_F] 
\end{array}
\end{equation}
For ease of notation, we denote
\small
\[\mathcal{G}(W, \alpha; \mathcal{D}^{\text{B-tr}}, W_0) = \mathcal{L}(g(W, \alpha, W_0); \mathcal{D}^{\text{B-tr}}) + \lambda_1 \norm{W}^2_F \] 
\normalsize
$W^{\prime}$ is plugged into the objective function of stage II. The gradient of the objective function in stage II with respect to $\alpha$ is computed to update $\alpha$:
\begin{equation}
\label{eqn:alpha}
\small
\begin{aligned}
\quad & \alpha^{*} \approx \alpha^{\prime} = \alpha - \eta_{\alpha}\nabla_{\alpha} [\mathcal{L}(g(W^\prime, \alpha, W_0); \mathcal{D}^{\text{B-val}}) \\& \quad \quad + \lambda_2 \norm{\alpha}^2_F]
\end{aligned}
\end{equation}
\begin{algorithm}[t]
	\caption{Optimization algorithm}
     Training dataset - $\mathcal{D}^{\text{tr}}$, BLO training dataset - $\mathcal{D}^{\text{B-tr}}$, BLO validation dataset - $\mathcal{D}^{\text{B-val}}$.
	\begin{algorithmic}
    \State \# Search phase
		\While {not converged}
		\State Update task weights $W$ using Equation (\ref{eqn:W}) on $\mathcal{D}^{\text{B-tr}}$
		\State Update attention parameter $\alpha$ using Equation (\ref{eqn:alpha}) on $\mathcal{D}^{\text{B-val}}$
		\EndWhile \\
    \State \# Finetune phase
    \State Using the learned $\alpha^{\prime}$, finetune the task weights further on $\mathcal{D}^{\text{tr}}$ until convergence.
	\end{algorithmic}
	\label{optim_algorithm} 
\end{algorithm}
Equation~\ref{eqn:alpha} can be further reduced as follows.  The chain rule is applied to estimate the gradient of the loss function in stage II with respect to $\alpha$. $W^{\prime}$ is an implicit on $\alpha$ as discussed above. The first part of the gradients can be decomposed as follows,

\small
\begin{equation}
\label{eqn:del_alpha}
\begin{array}{ll}
\nabla_{\alpha} \mathcal{L}(g(W^\prime, \alpha, W_0); \mathcal{D}^{\text{B-val}}) \nonumber \\ = \nabla_{\alpha} \mathcal{L}(g(W - \eta_{w}\nabla_{W} \mathcal{G}(W, \alpha; \mathcal{D}^{\text{B-tr}}, W_0), \alpha, W_0); \mathcal{D}^{\text{B-val}}) \nonumber\\
= \nabla_{\alpha} \mathcal{L}(g(W^\prime, \alpha, W_0); \mathcal{D}^{\text{B-val}}) -\eta_{w} \times \nonumber \\
\nabla^2_{\alpha,W} \mathcal{G}(W, \alpha;\mathcal{D}^{\text{B-tr}}, W_0)\nabla_{W^{\prime}} \mathcal{L}(g(W^\prime, \alpha, W_0); \mathcal{D}^{\text{B-val}}) 
\end{array}
\end{equation}
\normalsize
The above gradient estimation contains an expensive matrix-vector product that can be reduced using finite difference approximation:
\begin{equation}
\label{eqn:finite_diff}
\small
\begin{aligned}
\nabla^2_{\alpha,W} \mathcal{G}(W, \alpha;\mathcal{D}^{\text{B-tr}}, W_0)\nabla_{W^{\prime}}  \mathcal{L}(W^\prime, \alpha;\mathcal{D}^{\text{B-val}}, W_0) = \\ \frac{\nabla_{\alpha}  \mathcal{G}(W^+, \alpha;\mathcal{D}^{\text{B-tr}}, W_0) - \nabla_{\alpha} \mathcal{G}(W^-, \alpha;\mathcal{D}^{\text{B-tr}}, W_0)}{2 \epsilon }, \nonumber
\end{aligned}
\end{equation}
where
\begin{equation}
\small
\begin{aligned}
     W^{\pm} &= W \pm \epsilon{\nabla}_{W^{\prime}}  \mathcal{L}(W^\prime, \alpha;\mathcal{D}^{\text{B-val}}, W_0), \nonumber \\
     \epsilon &= \frac{0.01}{{\lVert \nabla_{W^{\prime}}  \mathcal{L}(W^\prime, \alpha;\mathcal{D}^{\text{B-val}}, W_0) \rVert}_2}. \nonumber 
\end{aligned}
\end{equation}
The algorithm is run iteratively until convergence to estimate $\alpha^* \approx \alpha^{\prime}$. Finetuning phase is further conducted on $\mathcal{D}^{\text{tr}}$ with the learned attention parameter $\alpha^{\prime}$. 

We analyze the loss curves of both the inner and outer optimization processes in our algorithm to determine the optimal hyperparameter choices. Given that our task involves solving a bilevel optimization problem, and considering the significance of the model's performance on \(\mathcal{D}^{\text{B-val}}\) (which pertains to the outer problem), we regard the convergence of the outer loss as a crucial criterion for overall convergence.

\begin{table*}[t]
\centering 
\small
\begin{tabular}{l | c | c | c | c | c | c | c | c} 
\toprule
\textbf{Dataset} & \textbf{RTE} & \textbf{MRPC} & \textbf{STS-B} & \textbf{CoLA} & \textbf{SST-2} & \textbf{QNLI} & \textbf{QQP} & \textbf{MNLI} \\
\hline
\textbf{Train Examples} & 2.5k & 3.7k & 5.7k & 8.5k & 67k & 105k & 364k & 393k \\
\textbf{Dev Examples} & 277 & 408 & 1.5k & 1.0k & 872 & 5.5k & 40k & 4.8k \\
\hline
\textbf{Metrics} & Acc & F1 & SCC & MCC & Acc & Acc & Acc & Acc \\
\bottomrule
\end{tabular}
\caption{The table details the eight datasets utilized in this study, sourced from the GLUE benchmark. Here, \textit{Acc} denotes Accuracy, \textit{SCC} refers to the Spearman Correlation Coefficient, and \textit{MCC} signifies the Matthews Correlation Coefficient.}
\label{tab:GLUE_datasets}
\end{table*}

\begin{table*}[t]
\centering 
\small
\begin{tabular}{l | c | c | c | c | c | c } 
\toprule
\textbf{Models} & \textbf{Datasets} & \textbf{Batch Size} & \textbf{Learning Rate} & \textbf{Epochs/Steps} & \textbf{Warmup Ratio/Steps} & \textbf{Weight Decay} \\
\hline
\multirow{1}{*}{BERT} & all & 16 &2e-5  & 3 epochs  & 10\%  & 0.01 \\
\hline
\multirow{4}{*}{RoBERTa} & RTE &  16 & 2e-5 & 2036 steps & 122 steps & 0.1\\
& MPRC &  16 & 1e-5 & 2296 steps & 137 steps & 0.1\\
& STS-B &  16 & 2e-5 & 3598 steps & 214 steps & 0.1\\
& CoLA &  16 & 1e-5 & 5336 steps & 320 steps & 0.1\\
\hline
\multirow{4}{*}{DeBERTa} & RTE &  32 & 1e-5 & 6 epochs & 50 steps & 0.01\\
& MPRC &  32 & 1e-5 & 6 epochs & 50 steps & 0.01\\
& STS-B &  32 & 7e-6 & 4 epochs & 100 steps & 0.01\\
& CoLA &  32 & 7e-6 & 6 epochs & 100 steps & 0.01\\
\hline
\multirow{4}{*}{BART} & RTE &  32 & 1e-5 & 1018 steps & 61 steps & 0.01\\
& MPRC &  64 & 2e-5 & 1148 steps & 68 steps & 0.01\\
& STS-B &  32 & 2e-5 & 1799 steps & 107 steps & 0.01\\
& CoLA &  64 & 2e-5 & 1334 steps & 80 steps & 0.01\\
\hline
\multirow{4}{*}{XLNet} & RTE &  32 & 2e-5 & 3000 steps & 500 steps & 0.01\\
& MPRC &  32 & 2e-5 & 800 steps & 200 steps & 0.01\\
& STS-B &  32 & 2e-5 & 800 steps & 200 steps & 0.01\\
& CoLA &  32 & 2e-5 & 3000 steps & 500 steps & 0.01\\
\bottomrule
\end{tabular}

\caption{Hyperparameter settings, as reported in their official repository, for all the different PLMs used in this work. For the learning rates, as an example, 2e-5 means $2\times 10^{-5}$.}
\label{tab:hyperparams}
\end{table*}

\begin{table*}[t]
\centering 
\small

\subfloat[Results for 300 training samples.]{
\centering
\begin{tabular}{l | c c c c} 
\toprule
\textbf{Datasets} & \textbf{Vanilla} & \textbf{DPS Dense} & \textbf{Child Tuning} & \textbf{Ours} \\
\midrule
\textbf{CoLA}  & 24.07 $\pm$ 17.04 & 23.85 $\pm$ 16.12 & 23.47 $\pm$ 14.48 & \textbf{41.81} $\pm$ 5.57 \\
\textbf{RTE}   & 58.09 $\pm$ 2.64  & 55.19 $\pm$ 4.54  & 58.77 $\pm$ 3.07  & \textbf{60.54} $\pm$ 2.59 \\
\textbf{STSB}  & 78.08 $\pm$ 3.21  & 74.07 $\pm$ 5.94  & 79.62 $\pm$ 2.14  & \textbf{82.09} $\pm$ 4.88 \\
\textbf{SST-2} & 80.52 $\pm$ 12.87 & 82.79 $\pm$ 4.06  & 77.67 $\pm$ 7.63  & \textbf{89.52} $\pm$ 0.78 \\
\textbf{MRPC}  & 80.57 $\pm$ 1.23  & \textbf{81.66} $\pm$ 0.53  & 81.09 $\pm$ 1.25  & {80.63} $\pm$ 1.47 \\
\textbf{QQP}   & 67.73 $\pm$ 3.95  & 67.14 $\pm$ 3.59  & 68.45 $\pm$ 4.46  & \textbf{72.54} $\pm$ 1.67 \\
\textbf{QNLI}  & 71.03 $\pm$ 8.92  & 68.79 $\pm$ 7.79  & 70.84 $\pm$ 8.42  & \textbf{76.16} $\pm$ 2.77 \\
\textbf{MNLI}  & 40.19 $\pm$ 2.67  & 40.04 $\pm$ 2.42  & 39.85 $\pm$ 2.55  & \textbf{48.44} $\pm$ 5.01 \\
\hline
\textbf{AVG}   & 62.54 $\pm$ 6.57  & 61.69 $\pm$ 5.62  & 62.47 $\pm$ 5.5   & \textbf{68.97} $\pm$ 3.09 \\
\bottomrule
\end{tabular}%
\label{tab:low-resource-300}
}

\vspace{0.1cm}

\subfloat[Results for 500 training samples.]{
\centering
\begin{tabular}{l | c c c c} 
\toprule
\textbf{Datasets} & \textbf{Vanilla} & \textbf{DPS Dense} & \textbf{Child Tuning} & \textbf{Ours} \\
\midrule
\textbf{CoLA} & 26.01 $\pm$ 13.2 & {{39.42 $\pm$ 11.8}} & 38.15 $\pm$ 14.10 & \textbf{49.35} $\pm$ 1.71\\
\textbf{RTE} & 58.30 $\pm$ 4.90 & 58.34 $\pm$ {2.90} & {{59.78}} $\pm$ {{5.02}} & \textbf{62.96} $\pm$ {{4.40}} \\
\textbf{STSB} & 81.77 $\pm$ 2.69 & {{83.38 $\pm$ 1.55}} & 80.87 $\pm$ 3.97 & \textbf{86.85} $\pm$ 0.52 \\
\textbf{SST-2} & 86.51 $\pm$ 6.48 & \textbf{89.38} $\pm$ 0.52 & {{89.04}} $\pm$ 1.34 & 88.98 $\pm$ {{0.91}} \\
\textbf{MRPC} & 82.96 $\pm$ {{0.84}} & \textbf{83.29} $\pm$ 0.78 & 81.42 $\pm$ 1.88 & {{83.21}} $\pm$ 2.30 \\
\textbf{QQP} & 71.68 $\pm$ 1.90 & 74.43 $\pm$ {{1.35}} & {{74.53}} $\pm$ 1.45 & \textbf{74.80} $\pm$ 0.58 \\
\textbf{QNLI} & 76.82 $\pm$ 2.09 & {{77.09}} $\pm$ {1.75} & 76.73 $\pm$ 3.19 & \textbf{79.78} $\pm$ {{1.83}} \\
\textbf{MNLI} & 42.81 $\pm$ 4.49 & {{46.61}} $\pm$ {2.70} & 46.33 $\pm$ {{3.96}} & \textbf{53.41} $\pm$ {4.84} \\ \hline
\textbf{AVG} & 65.85 $\pm$ 4.57 & {{68.99}} $\pm$ {{2.92}} & 68.35 $\pm$ 4.36 & \textbf{72.42} $\pm$ 2.14 \\
\bottomrule
\end{tabular}%
\label{tab:low-resource-500}
}

\vspace{0.1cm}

\subfloat[Results for 1000 training samples.]{
\centering
\begin{tabular}{l | c c c c} 
\toprule
\textbf{Datasets} & \textbf{Vanilla} & \textbf{DPS Dense} & \textbf{Child Tuning} & \textbf{Ours} \\
\midrule
\textbf{CoLA} & 47.97 $\pm$ 5.62 & {{52.89}} $\pm$ {{2.50}} & 51.69 $\pm$ 2.81 & \textbf{54.19} $\pm$ 1.77 \\
\textbf{RTE} & 62.60 $\pm$ 3.46 & {{64.44}} $\pm$ {1.76} & 63.93 $\pm$ 5.15 & \textbf{67.62} $\pm$ {{2.53}} \\
\textbf{STSB} & 85.86 $\pm$ 1.34 & {{87.15}} $\pm$ 1.77 & 87.08 $\pm$ {{0.79}} & \textbf{88.31} $\pm$ 0.78 \\
\textbf{SST-2} & 90.28 $\pm$ 0.55 & \textbf{90.92} $\pm$ 0.64 & 89.92 $\pm$ 2.95 & {{90.46}} $\pm$ {{0.66}} \\
\textbf{MRPC} & 85.34 $\pm$ 1.30 & {{85.90}} $\pm$ {0.86} & 84.81 $\pm$ 1.81 & \textbf{87.03} $\pm$ {{1.76}} \\
\textbf{QQP} & 76.96 $\pm$ 1.50 & {77.81} $\pm$ 0.57 & 76.22 $\pm$ 2.60 & \textbf{77.81} $\pm$ {{0.67}} \\
\textbf{QNLI} & 81.61 $\pm$ 1.32 & {{82.58}} $\pm$ 1.19 & 82.42 $\pm$ {{1.18}} & \textbf{82.94} $\pm$ 1.02 \\
\textbf{MNLI} & 54.93 $\pm$ 5.94 & {{58.33 $\pm$ 3.56}} & 56.56 $\pm$ 4.99 & \textbf{65.07} $\pm$ 3.48 \\ \hline
\textbf{AVG} & 73.19 $\pm$ 2.62 & {{75.00 $\pm$ 1.61}} & 74.07 $\pm$ 2.75 & \textbf{76.68} $\pm$ 1.58\\
\bottomrule
\end{tabular}%
\label{tab:low-resource-1000}
}

\caption{We present a comparison of our method, Vanilla, DPS dense, and $\text{CHILD-TUNING}_D$ method on $\text{BERT}_{\text{LARGE}}$~\cite{lee2019mixout} across various low-resource scenarios. We report the mean and standard deviation of ten random seeds.
\textbf{Bold} indicates the best performance. On average, our method outperforms the best baseline by 6.43 points in the setting with 300 training samples, by 3.43 points in the setting with 500 training samples, and outperforms it by 1.68 points for 1000 training samples.}
\label{tab:low-resource-300-500-1000}
\end{table*}

\section{Datasets}\label{sec:appendix B.}
We evaluated our method on various datasets from the GLUE benchmark~\cite{warstadt2018neural, wang-etal-2018-glue}. In this section, we describe each of the datasets used in this work. Table~\ref{tab:GLUE_datasets} summarizes the split statistics and the evaluation metric for each dataset. The CoLA dataset~\cite{warstadt2018neural}, known as the Corpus of Linguistic Acceptability, serves as a benchmark to assess the natural language processing model's capacity to comprehend and predict the acceptability of English sentences. On the other hand, the RTE dataset~\cite{dagan2006pascal} focuses on determining the logical entailment between sentences. QNLI dataset~\cite{devlin2018bert} originates from the Stanford Question Answering Dataset (SQuAD) and aims to evaluate the model's competence in comprehending and inferring information from questions and corresponding contextual paragraphs. MNLI dataset's objective involves classifying sentence relationships as entailment, contradiction, or neutral, and it encompasses diverse genres and domains, providing a more comprehensive and challenging assessment of models' inference abilities in varied contexts~\cite{williams2018broad}. Lastly, MRPC~\cite{dolan2005automatically}, STS-B~\cite{cer2017semeval}, and QQP datasets~\cite{wang-etal-2018-glue} are employed to tackle paraphrase identification, semantic textual similarity, and question pair similarity tasks, respectively. The Stanford Sentiment Treebank (SST-2) dataset~\cite{socher2013recursive} is a sentiment classification task on sentences from movie reviews.

\section{More about related works} \label{sec:appendix C.}
\paragraph{Prior finetuning regularization-based approaches} We compare our method with several baselines on $\text{BERT}_{\text{LARGE}}$ model~\cite{devlin2018bert}. It's worthwhile to note that this is a challenging task. Various methods are proposed to tackle this challenge. Mixout~\cite{lee2019mixout} stochastically replaces model parameters with pretrained parameters (with probability $p$) to mitigate catastrophic forgetting issue. R3F \cite{aghajanyan2020better} introduces parametric noise into input sentence embeddings for robustness. R-Dropout~\cite{wu2021r} mitigates the two-directional KL divergence by assessing the output distributions of two submodels, each independently generated via Dropout. In Re-init \cite{zhang2020revisiting}, the pooler and top $K$ BERT transformer layers are reinitialized from the distribution~$\mathcal{N}(0, {0.02^2})$, which is the original BERT initialization. ${\text{CHILD-TUNING}_D}$~\cite{xu2021raise} proposes to finetune a child network selected using FIM, constructed from the gradients of the training dataset, with the non-child network frozen to pretrained weights. DPS dense~\cite{zhang2022fine} provides an alternative technique for selecting child networks, utilizing a two-stage update algorithm. The initial stage involves the computation of a Gradient Accumulation Matrix (GAM), which estimates empirical FIM, for all parameters within the model. Following that, in the second stage, a child network is selected according to the GAM, after which its corresponding weights undergo modification. 

Very recent work by \citet{tong2023bi} has introduced a novel approach to select optimal sub-networks. Drawing inspiration from \citet{wu2021r}, their method selects sub-networks leveraging gradient from sub-models generated by dropout, sidestepping the Fisher information matrix. Nevertheless, they still retain a thresholding approach for sub-network selection based on mini-batch gradients, akin to FIM-based approaches, which is sub-optimal.
In contrast, our methodology diverges from a discrete sub-network selection, opting for a bi-level continuous optimization strategy tailored to improve performance on a downstream task. 
Since their code is not available in their linked GitHub repository, by comparing our results against those of \citet{tong2023bi} directly, the benefits of our methodology become evident. Evaluating on various settings using the $\text{BERT}_{\text{LARGE}}$ model, we observed the following: On low-resource settings, for 500 samples, our method scores an average of 72.42, outperforming their 71.06 (as seen in Table~\ref{tab:low-resource}); With 1000 samples, we achieve an average score of 76.68, superior to their 76.21 (Table~\ref{tab:low-resource}). On few-sample BERT finetuning, our average score stands at 80.42 compared to their 80.26 (Table~\ref{tab:mainresult}).

\paragraph{Weight space manipulations} Task vectors in weight space \cite{ilharco2022editing}, which represent the difference between task and pretrained weights in a model, can be manipulated through arithmetic operations to guide a model towards specific tasks. \citet{ilharco2022patching} introduces PAINT, a model patching method for open-vocabulary models like CLIP, which employs a convex combination of task and pretrained weights to adapt to new tasks to be patched while preserving accuracy on previous tasks, the mixing coefficient is manually assigned. In contrast to these works, our approach presents a regularization technique where resultant weights are represented as a weighted summation of task and pretrained weights controlled by the attention parameter, specifically aiming to mitigate issues in finetuning pretrained language models on small downstream datasets. We learn the attention parameters and task weights using BLO framework.

\section{Hyperparameters settings} \label{sec:appendix D.}

We evaluate our method on different PLM's including $\text{BERT}_{\text{LARGE}}$~\cite{devlin2018bert}, $\text{RoBERTa}_{\text{LARGE}}$~\cite{liu2019roberta}, $\text{BART}_{\text{LARGE}}$~\cite{lewis2019bart}, $\text{DeBERTa}_{\text{LARGE}}$~\cite{he2020deberta}, and
$\text{XLNet}_{\text{LARGE}}$~\cite{yang2019xlnet}. 

For the experimental results presented in Sections~\ref{sec:Main}, \ref{sec:low_resource}, \ref{sec:sufficient_samples}, and \ref{sec:ablation_studies}, we employed the $\text{BERT}_{\text{LARGE}}$ model, following~\citet{xu2021raise,zhang2022fine}. Our algorithm consists of two phases, 1) Search phase and 2) Finetune phase. We report the hyperparameters used for each phase and grid search for the best hyperparameter settings for a task on a split of the training set following~\citet{xu2021raise,zhang2022fine} for the $\text{BERT}_{\text{LARGE}}$ model.

Consistent with prior works in low-resource settings, such as those in our baseline methods~\cite{xu2021raise,zhang2022fine}, we randomly select a subset of the specified size (300, 500, or 1000) from the entire dataset based on a seed. This approach is commonly adopted in this field to avoid bias towards any specific fixed subset. We then report the mean and standard deviation of the evaluation metric across multiple seeds for all methods.

We partition $\mathcal{D}^{\text{tr}}$ into two subsets of $\mathcal{D}^{\text{B-tr}}$ and $\mathcal{D}^{\text{B-val}}$. We conducted experiments with different data ratios for $\mathcal{D}^{\text{B-tr}}$ and $\mathcal{D}^{\text{B-val}}$, including 50\%-50\% and 80\%-20\%. Empirically, we observed that the 80\%-20\% split performs better in low-resource settings compared to a 50\%-50\% split. This is likely because allocating more training data points to the inner optimization of task weights is advantageous, especially given the greater number of parameters (task weights) at the inner optimization level.

\subsection{Search phase}
The weights mixup is applied to all the parameters with ``nn.Linear layers''. The learning rate of $W$ is set to $2\times 10^{-5}$. The weight decay is set to 0.01. The warmup ratio is 10\%. The batch size is set to 16. AdamW~\cite{loshchilov2017decoupled} optimizer is used for $W$. We initialize the $\alpha$ values by sampling the entries of $\alpha_1$ and $\alpha_2$ from a Gaussian distribution with mean of 1 and standard deviation of 0.005. The learning rate of $\alpha$ is set to $2\times 10^{-3}$. In this work, the warmup ratio for $\alpha$ is set equal to the warmup ratio of $W$ which is 10\%. We use an AdamW optimizer with $\beta_1$ = 0.9, $\beta_2$ = 0.999, $\epsilon$ = $ 10^{-8}$, and weight decay of 0.01. Since it is an optimization problem, we search for the hyperparameters that lead to a smooth reduction in the loss curves. We grid search for the number of epochs in \{1,2,3,5\}. The maximum sequence length is 128. We partition the original training dataset, represented as $\mathcal{D}^{\text{tr}}$, into two subsets of 80\% and 20\% split randomly to perform the search phase. To further mitigate overfitting, we perform this random sampling $K$ times to produce $K$ different sets of $\mathcal{D}^{\text{B-tr}}$ and $\mathcal{D}^{\text{B-val}}$. 

We then perform the search phase $K$ times to get $K$ different learned parameters. We average them to obtain optimal learned parameters of the search phase. We search in $K$ = \{1,2,5\}. Though $K$ = 5 gave best results, $K$ = 2 saves computation time and gives slightly less but comparative results. For instance, in the evaluation of the CoLA dataset presented in Table~\ref{tab:mainresult}, setting K to 5 yielded an average score of 66.07, whereas setting K to 2 resulted in a score of 65.10. Both configurations outperformed the vanilla model, which achieved a score of 64.11.

We use a low-rank approximation of $\alpha$ for estimating
the child net structure. Since we are mainly targeting the low-resource domain, 
the low-rank approximation of $\alpha$ mitigates overfitting. In our experimental setup, we consider the rank values of \{1, 2, 8\} for the dimensionality reduction of $\alpha$, selecting the best performing value empirically. We found that a rank of $r$ = 1 consistently produced superior results, thus, this rank is adopted across all experimental trials. We didn't try an even larger rank or full rank of $\alpha$ as a full rank of $\alpha$ is also computationally heavy to perform. 

We initialize the $\alpha_1$ and $\alpha_2$ parameters using a Gaussian distribution. The mean of this distribution is set to 1, and we use a standard deviation of 0.005. We clip the initial alphas to [0, 1] to ensure they are valid attention parameters. This initialization strategy is designed to provide a balanced starting point for the optimization process. 

In addition, we have conducted experiments on our method both with and without weight decay for $\alpha$ parameters. Based on our empirical findings, we observed that employing weight decay yields better results compared to not using it. The likely reason for this improvement is that weight decay encourages the $\alpha$ values to be closer to 0. This inclination towards lower values results in a higher contribution from the pretrained weights in the resultant weight estimation, which in turn aids in regularizing the network. Such regularization appears to be beneficial for the overall performance and stability of the model in our experiments.

\subsection{Finetune phase}
With the optimal $\alpha^*$ in search phase, we further finetune for $W$ on entire training dataset. We use the default hyperparameter settings for $W$ following~\citet{devlin2018bert}. The number of epochs is searched in \{1,3\} and learning rate is searched in $\{2\times 10^{-5}, 3\times 10^{-6}\}$.

For different PLMs in Section~\ref{sec:diff_plms}, we use the default settings for $W$ in their official repository for both the search and the finetune phase, without exhaustive hyperparameter sweep. The number of training steps/epochs, warm-up steps and batch size for the finetuning phase are given in Table~\ref{tab:hyperparams}. For learning $\alpha$ in search phase, we use $\alpha$ learning rate $2\times 10^{-3}$, the $\alpha$ warmup ratio is set roughly the same as the warmup ratio of $W$ for each of the different PLM, i.e., warmup ratio for RoBERTa is 0.06, DeBERTa is 0.06, XLNet is 0.17, and BART is 0.1. For $\alpha$, we use an AdamW optimizer with $\beta_1$ = 0.9, $\beta_2$ = 0.999, $\epsilon = 10^{-8}$ and the weight decay is 0.01. 

\subsection{Baselines}
For the baselines, the following hyperparameter search space is used following their respective original papers: 
\begin{itemize}
    \item Mixout: Mixout probability $p \in$ \{0.1, 0.2, 0.3, 0.4, 0.5, 0.6, 0.7, 0.8\}.
    \item R3F: Noise types $\in \{\mathcal{N}, \mathcal{U}\}$, $\sigma \in \{10^{-5}\}$, $\lambda \in \{0.1, 0.5, 1, 5\}$.
    \item Re-init: L $\in$ \{1, 2, 3, 4, 5, 6, 7\}.
    \item ${\text{CHILD-TUNING}_D}$: $p_D \in \{0.1, 0.2, 0.3\}$, learning rate in $\{2\times 10^{-5}, 4\times 10^{-5}, 6\times 10^{-5}, 8\times 10^{-5}, 10^{-4}\}$.
    \item R-Dropout: $p \in$ \{0.1\} and $\alpha \in \{0.1, 0.5, 1, 3, 5\}$.
    \item DPS dense: $p \in$ \{0.1, 0.2, 0.3, 0.4, 0.5\} and update ratio $ur \in \{0.05, 0.1, 0.2\}$.
\end{itemize}
We use a single A100 GPU machine for our experiments. The mean and the standard deviation on the original development set over ten random seeds are reported following~\citet{xu2021raise, zhang2022fine}.



\section{Performance on low-resource scenarios} \label{sec:appendix G.}

Due to space constraints, we only present the average scores across datasets in Table~\ref{tab:low-resource}. The results on each of the eight datasets across 300, 500, 1K training data splits are presented in Table~\ref{tab:low-resource-300-500-1000}. We report both the mean and the standard deviation of the metric, evaluated on the test set for each dataset, across ten random seeds. On average, our method outperforms the best baseline by 6.43 points in the setting with 300 training samples, by 3.43 points in the setting with 500 training samples, and outperforms it by 1.68 points for 1000 training samples.

\section{Comparison with parameter efficient finetuning methods} \label{sec:peft_appendix.}

Similarly, due to space constraints, we only present the average scores across datasets in Figure~\ref{fig:peft_fig}. The results on each datasets across 500, 1K training data splits are presented in Table~\ref{tab:lora}. We report both the mean and the standard deviation of the metric, evaluated on the test set for each dataset, across ten random seeds. 


We implemented the PEFT baselines using the PEFT library~\cite{peft}. In conducting our hyperparameter search, we adhered to the hyperparameter settings provided in their respective original papers and the examples from the PEFT library. For LoRA, we set the rank $r$ to 8, the scaling factor $\alpha$ to 16, and included a dropout of 0.1. Consequently, LoRA has 788,482 trainable parameters (0.236\% of total). In the case of Prefix-Tuning, we set the number of virtual tokens to 20. It has 985,090 trainable parameters (0.294\% of total). For Prompt Tuning, we set the prompt length to 20. It has 22,530 trainable parameters (0.0067\% of the total). In terms of time complexity, assuming vanilla is x1, Prompt Tuning is ~0.2x, Prefix tuning is ~0.4x and LoRA is ~0.3x.

\section{Performance on sufficient samples}\label{sec:sufficient_samples}
\begin{figure}[t]
    \centering
    \includegraphics[width=\linewidth, height=0.2\textheight]{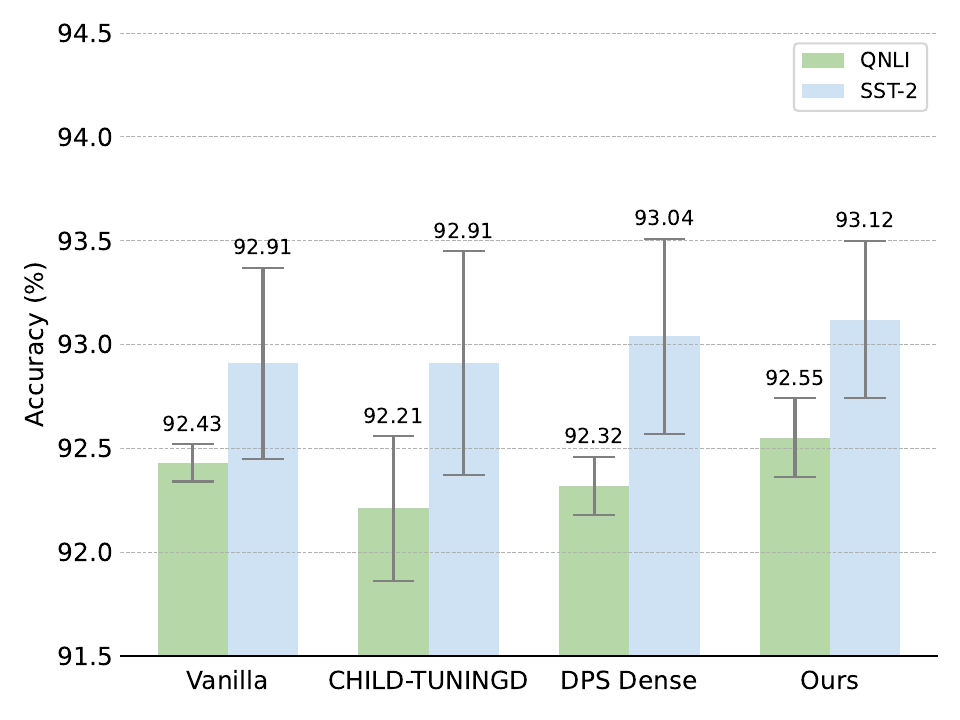}
    \caption{
    Comparison of our method with Vanilla, $\text{CHILD-TUNING}_D$, and DPS dense method on the QNLI and SST-2 datasets with sufficient training examples. The bar plots represent the mean accuracy from ten random seeds, and error bars denote the standard deviation.
    }
    \label{fig:combined_figure}
\end{figure}
            
            


     
We investigate whether our method can also improve accuracy on larger training datasets. To this end, we train $\text{BERT}_{\text{LARGE}}$ on SST-2 and QNLI datasets, both of which have over 50K training examples. We compare our method to the baselines $\text{CHILD-TUNING}_D$ and DPS dense. The results are presented in Fig~\ref{fig:combined_figure}, respectively. On the QNLI dataset, both $\text{CHILD-TUNING}_D$ and DPS dense perform worse than the vanilla method. However, our method yields a small accuracy improvement over the vanilla method. The small gaps between the performance of these methods are reasonable, because as the number of training instances increases, the impact of regularization reduces, leading to non-significant improvements with regularization-based methods in large sample settings. 

\section{Comparison with model soup}\label{sec:model_soup}
In this section, we compare our method against a baseline method inspired from the uniform model soup approach~\cite{wortsman2022model} in the low-resource settings. 
We average over five models, each trained using vanilla finetuning on the training dataset adhering to the optimal hyperparameter configuration as detailed in \citet{devlin2018bert} and initialized with a different random seed. The performance of this averaged model is then evaluated on an unseen test dataset. This entire process is repeated ten times to obtain the mean and the standard deviation.

Our observations, summarized in Table~\ref{tab:results_comparison_model_soup}, indicate that the model soup approach yields mixed results over vanilla finetuning in low-resource settings. 
Conversely, our method demonstrates consistent superiority over model soup. 
Our method leverages an attention-guided weight mixup, optimized through BLO across two distinct splits of the training dataset, resulting in more pronounced performance improvements.

\begin{table*}[t]
\centering 
\small

\subfloat[Results for 500 training samples.]{
\centering
\begin{tabular}{l | c c c c c} 
\toprule
\textbf{Datasets} & \textbf{Vanilla} & \textbf{Prompt Tuning} & \textbf{Prefix-Tuning} & \textbf{LoRA} & \textbf{Ours} \\
\midrule
\textbf{CoLA} & 26.01 $\pm$ 13.2 & 13.41 $\pm$ 9.95 & 42.96 $\pm$ 4.96 & 47.26 $\pm$ 2.57 & \textbf{49.35} $\pm$ 1.71 \\
\textbf{RTE} & 58.30 $\pm$ 4.90 & 56.28 $\pm$ 2.16 & 55.77 $\pm$ 2.63 & 57.15 $\pm$ 1.87 & \textbf{62.96} $\pm$ 4.40 \\
\textbf{STSB} & 81.77 $\pm$ 2.69 & 58.38 $\pm$ 16.52 & 74.44 $\pm$ 4.99 & 80.22 $\pm$ 2.93 & \textbf{86.85} $\pm$ 0.52 \\
\textbf{MRPC} & 82.96 $\pm$ 0.84 & 81.85 $\pm$ 0.3 & 78.63 $\pm$ 2.31 & 79.88 $\pm$ 1.53 & \textbf{83.21} $\pm$ 2.30 \\ 
\hline
\textbf{AVG} & 62.26 $\pm$ 5.41 & 52.48 $\pm$ 7.23 & 62.95 $\pm$ 3.72 & 66.13 $\pm$ 2.22 & \textbf{70.59} $\pm$ 2.23 \\
\bottomrule
\end{tabular}%
\label{tab:lora-500}
}

\vspace{0.1cm}

\subfloat[Results for 1000 training samples.]{
\centering
\begin{tabular}{l | c c c c c} 
\toprule
\textbf{Datasets} & \textbf{Vanilla} & \textbf{Prompt Tuning} & \textbf{Prefix-Tuning} & \textbf{LoRA} & \textbf{Ours} \\
\midrule
\textbf{CoLA} & 47.97 $\pm$ 5.62 & 23.69 $\pm$ 12.24 & 44.11 $\pm$ 14.79 & 51.46 $\pm$ 1.72 & \textbf{54.19} $\pm$ 1.77 \\
\textbf{RTE} & 62.60 $\pm$ 3.46 & 57.44 $\pm$ 0.91 & 59.21 $\pm$ 4.80  & 60.50 $\pm$ 2.79 & \textbf{67.62} $\pm$ 2.53 \\
\textbf{STSB} & 85.86 $\pm$ 1.34 & 69.29 $\pm$ 10.22 & 80.13 $\pm$ 1.58 & 83.57 $\pm$ 1.41 & \textbf{88.31} $\pm$ 0.78 \\
\textbf{MRPC} & 85.34 $\pm$ 1.30 & 81.99 $\pm$ 0.43 & 83.70 $\pm$ 2.09 & 83.98 $\pm$ 2.09 & \textbf{87.03} $\pm$ 1.76 \\ 
\hline
\textbf{AVG} & 70.44 $\pm$ 2.93 & 58.1 $\pm$ 5.95 & 66.79 $\pm$ 5.82 & 69.88 $\pm$ 2 & \textbf{74.29} $\pm$ 1.71 \\ 
\bottomrule
\end{tabular}%
\label{tab:lora-1000}
}

\caption{We present a comparison of our method with Vanilla, Prompt Tuning, Prefix-Tuning, and LoRA finetuning approaches on CoLA, RTE, STSB, and MRPC datasets in low-resource scenarios with 500 and 1000 training instances. We report both mean and standard deviation, each over ten runs. The \textbf{bold} values represent the highest performance in each scenario.}
\label{tab:lora}
\end{table*}

\begin{table*}[t]
\small
\centering
\begin{tabular}{c|ccc|ccc}
\hline
\multicolumn{1}{c|}{} & \multicolumn{3}{c|}{\textbf{500}} & \multicolumn{3}{c}{\textbf{1K}} \\ \hline
\multicolumn{1}{c|}{\textbf{Datasets}} & \textbf{Vanilla} & \textbf{Model Soup} & \textbf{Ours} & \textbf{Vanilla} & \textbf{Model Soup}& \textbf{Ours} \\ \hline
\textbf{CoLA} & 26.01 $\pm$ 13.2 & 26.33 $\pm$ 18.5 & \textbf{49.35} $\pm$ 1.71 & 47.97 $\pm$ 5.62 & 43.69 $\pm$ 12.51 & \textbf{54.19} $\pm$ 1.77 \\
\textbf{RTE} & 58.30 $\pm$ 4.90 & 58.56 $\pm$ 4.96 & \textbf{62.96} $\pm$ 4.40 & 62.60 $\pm$ 3.46 & 63.93 $\pm$ 5.15 & \textbf{67.62} $\pm$ 2.53 \\
\textbf{STSB} & 81.77 $\pm$ 2.69 & 82.75 $\pm$ 0.71 & \textbf{86.85} $\pm$ 0.52 & 85.86 $\pm$ 1.34 & 85.85 $\pm$ 0.72 & \textbf{88.31} $\pm$ 0.78 \\
\textbf{MRPC} & 82.96 $\pm$ 0.84 & 81.53 $\pm$ 0.84 & \textbf{83.21} $\pm$ 2.3 & 85.34 $\pm$ 1.3 & 83.40 $\pm$ 1.20 & \textbf{87.03} $\pm$ 1.76 \\
\bottomrule
\end{tabular}%
\caption{We present a comparison of our method, Vanilla, and model soup on RTE, MRPC, STSB, and CoLA. The reported results are mean and standard deviation of the evaluation metrics over ten random seeds for each dataset at the 500 and 1K training instances. \textbf{Bold} indicates the highest performance in each row.}
\label{tab:results_comparison_model_soup}
\end{table*}

\setlength{\tabcolsep}{4pt} 
\begin{table*}[t]
\centering 
\small
\begin{tabular}{l | c c | c c | c c | c c | c c | c c} 
\hline
\textbf{Datasets} & \multicolumn{2}{c|}{\textbf{Ours}} & \multicolumn{2}{c|}{\textbf{Vanilla}} & \multicolumn{2}{c|}{\textbf{ \makecell{Joint\\Training} }}  & \multicolumn{6}{c}{\textbf{$\text{Random}_{\alpha}$}}  \\
\cline{8-13} & & & & & & & \multicolumn{2}{c|}{ $\sigma_{\alpha}$ = 0.005} & \multicolumn{2}{c|}{$\sigma_{\alpha}$ = 0.1} & \multicolumn{2}{c}{$\sigma_{\alpha}$ = 0.45}\\
\cline{2-13} & mean & std & mean & std & mean & std & mean & std & mean & std & mean & std \\ 
\hline
\textbf{CoLA} & \textbf{66.07} & {1.35} & 64.11 & {1.33} & 64.18 & 1.19 & 64.03 & 1.96 & 64.37 & {1.00} & 47.52 & 16.02\\
\textbf{MRPC} & \textbf{91.84} & {0.37} & 90.80 & 1.77 & 89.77 & 2.91 & 91.37 & 0.66 & 90.65 & 0.82 & 85.42 & 2.94\\
\textbf{RTE} & \textbf{73.43} & {1.52} & 70.69 & 2.83 & 71.48 & 1.39 & 72.20 & {0.97} & 68.30 & 6.76 & 55.85 & 2.64\\
\textbf{STSB} & \textbf{90.34} & {0.48} & 89.92 & 0.61 & 90.03 & 0.42 & 89.85 & 0.54 & 89.97 & 0.52 & 88.38 & {0.18}\\
\hline
\textbf{AVG} & \textbf{80.42} & {0.93} & 78.88 & 1.64 & 78.86 & 1.48 & 79.36 & 1.03 & 78.32 & 2.27 & 69.29 & 5.44\\
\bottomrule
\end{tabular}

\caption{Comparision of our method with Vanilla, Joint-training and $\text{Random}_{\alpha}$ by varying $\sigma_{\alpha}$ 
to understand the effectiveness of the attention-based weights mixup and the need to learn $\alpha$ using a BLO framework.}
\label{tab:random_init}
\end{table*}

\end{document}